\documentclass[conference]{IEEEtran}
\IEEEoverridecommandlockouts

\usepackage{amsmath,amssymb,amsfonts}
\usepackage{algorithmic}
\usepackage{graphicx}
\usepackage{textcomp}
\usepackage{xcolor}
\usepackage{enumitem}
\usepackage{array,booktabs,tabularx}
\usepackage{wrapfig}
\usepackage[font={small,it}]{caption}
\usepackage{subcaption}
\usepackage{listings}
\usepackage{hyperref}
\usepackage[backend=bibtex,style=ieee]{biblatex}
\def\BibTeX{{\rm B\kern-.05em{\sc i\kern-.025em b}\kern-.08em
    T\kern-.1667em\lower.7ex\hbox{E}\kern-.125emX}}
\begin{document}
\title{Road Segmentation for ADAS/AD Applications\\
}
\author{
    \IEEEauthorblockN{Mathanesh Vellingiri Ramasamy}
    \IEEEauthorblockA{
        \textit{MSc Systems, Control and Mechatronics} \\
        \textit{Chalmers University of Technology} \\
        Gothenburg, Sweden \\
        Email: \href{mailto:matvell@chalmers.se}{matvell@chalmers.se}} \and 
\IEEEauthorblockN{Dimas Rizky Kurniasalim}
    \IEEEauthorblockA{
        \textit{Computer Science and Engineering} \\
        \textit{Chalmers University of Technology} \\
        Gothenburg, Sweden \\
        Email: \href{mailto:rizky@chalmers.se}
        {rizky@chalmers.se}}
}
      
\maketitle
\begin{abstract}
\noindent Accurate road segmentation is essential for autonomous driving and ADAS, enabling effective navigation in complex environments. This study examines how model architecture and dataset choice affect segmentation by training a modified VGG-16 on the Comma10k dataset and a modified U-Net on the KITTI Road dataset. Both models achieved high accuracy, with cross-dataset testing showing VGG-16 outperforming U-Net, despite U-Net being trained for more epochs. We analyze model performance using metrics such as F1-score, mIoU, and precision, discussing how architecture and dataset impact results.
\end{abstract}
\vspace{2mm}
\section{Introduction} 
\noindent Road image segmentation plays a crucial role in applications such as autonomous driving (AD), advanced driver assistance systems (ADAS), traffic monitoring, and smart city development. It enables a deeper understanding of the driving environment, supporting safer and more efficient decision-making. This is particularly valuable for tasks like lane detection, obstacle avoidance, and path planning in self-driving vehicles.\\

\noindent This report focuses on training two models on separate datasets, then performing cross-validation by evaluating one model with the other dataset it has not been trained on. For this, we trained a modified VGG-16 model with the Comma10k dataset and a modified U-Net model on the KITTI Road dataset. We will also investigate what went wrong with the under-performing model, and how to improve model training moving forward.

\section{Background Theory}

\subsection{Semantic Segmentation}
\noindent Semantic segmentation plays a crucial role in enabling machines to understand visual data at a granular level. Unlike image classification, which assigns a single label to an entire image, and object detection, which identifies objects within bounding boxes, semantic segmentation provides a dense prediction, assigning each pixel a label corresponding to a specific class (e.g., car, road, tree, sky). This pixel-level understanding is essential in applications where precise spatial information is required, such as AD/ADAS systems where safety is of upmost importance.

\begin{figure}[ht]
  \centering
  \fbox{\includegraphics[width=0.45\textwidth]{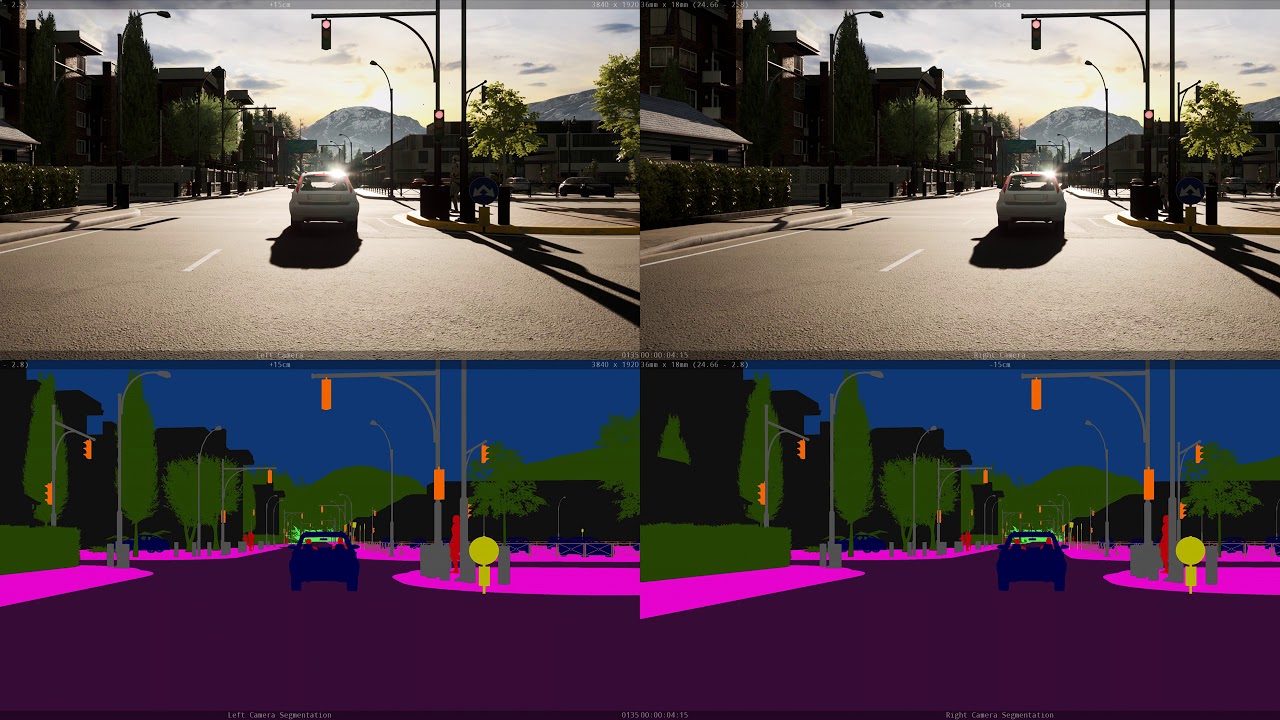}}
  \caption{An example of image segmentation with multiple classes, each class represented by their respective colors.}
  \label{fig:exampleseg}
\end{figure}

\subsection{Encoder-Decoder Networks}
\noindent Encoder-decoder architectures are foundational approaches in segmentation. The encoder compresses the input image to a lower-dimensional representation, capturing essential features while reducing spatial resolution. The decoder then reconstructs the image to its original resolution, using learned features to produce a dense, segmented output. 

\subsection{VGG-16 and U-Net}
\begin{figure}[ht]
  \centering
  \fbox{\includegraphics[width=0.4\textwidth]{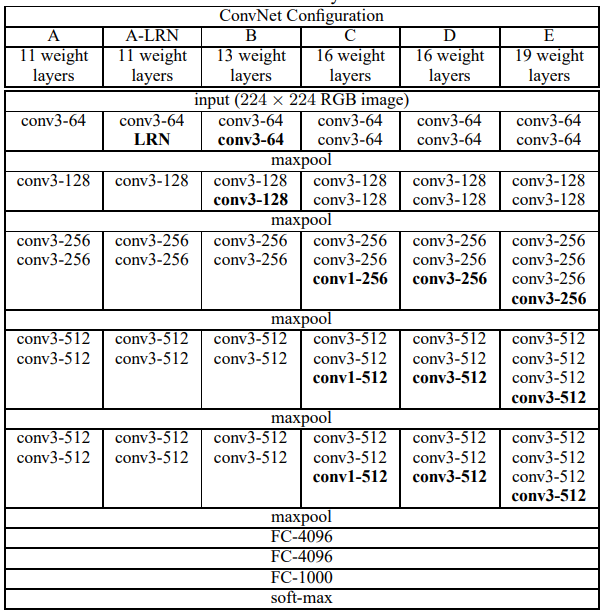}}
  \caption{The VGG-16 backbone model used without modification. Courtesy of Very Deep Convolutional Networks for Large-Scale Image Recognition by Karen Simonyan et al.}
  \label{fig:vgg_model}
\end{figure}

\begin{figure}[ht]
  \centering
  \fbox{\includegraphics[width=0.4\textwidth]{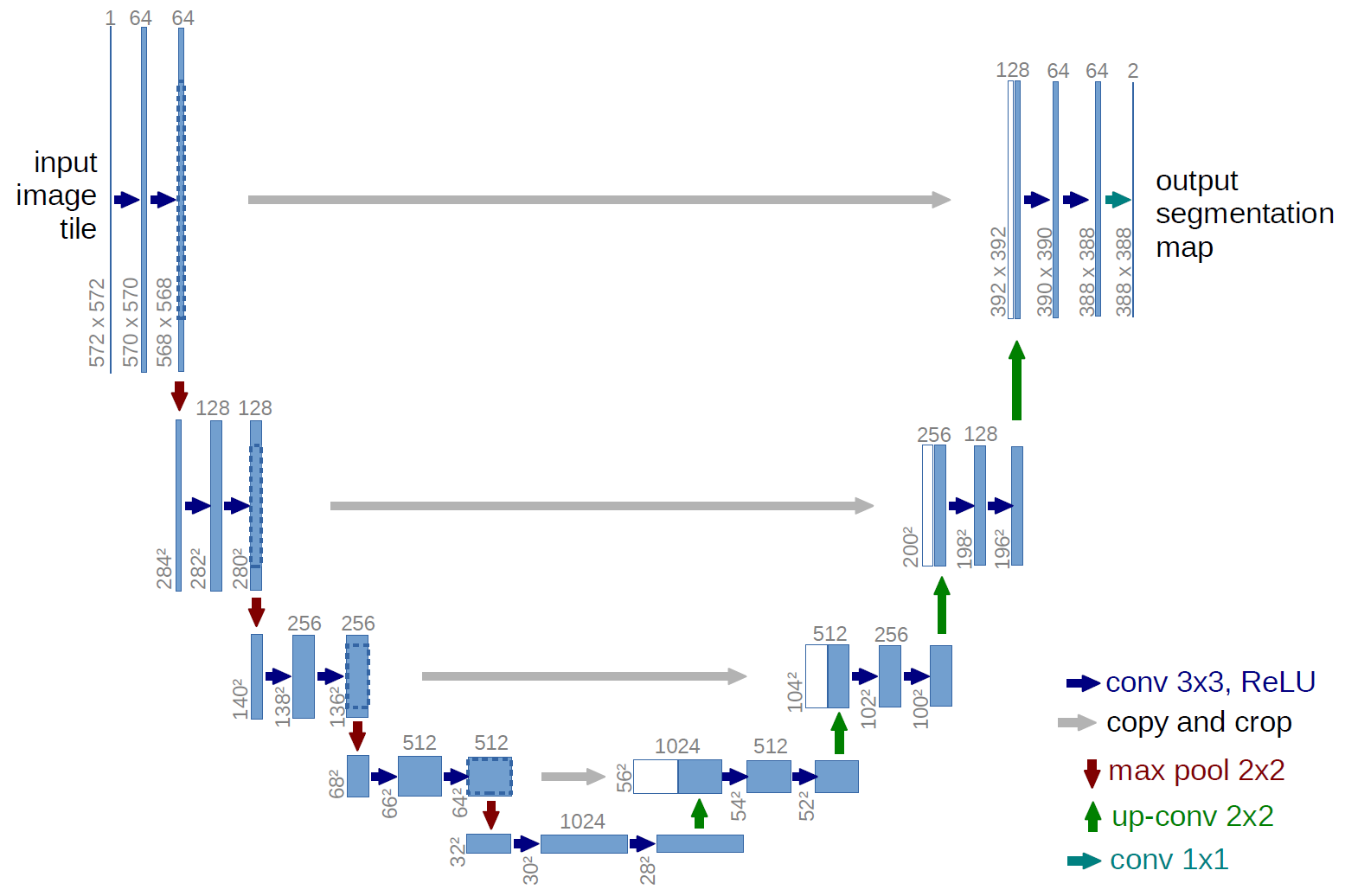}}
  \caption{The U-Net backbone model used without modification. Courtesy of U-Net: Convolutional Networks for Biomedical
Image Segmentation by Olaf Ronneberger et al.}
\label{fig:unet_model}
\end{figure}

\noindent VGG-16 and U-Net are two influential deep learning architectures widely used in image processing tasks. VGG-16, a convolutional neural network (CNN) developed by the Visual Geometry Group (VGG), is known for its 16-layer depth and sequential arrangement of small 3x3 convolutional filters, which allows it to capture intricate features in image classification tasks. In contrast, U-Net is a CNN architecture primarily designed for image segmentation, with a U-shaped structure that includes an encoder-decoder format. Its downsampling (encoder) path extracts spatial features, while the upsampling (decoder) path, enhanced by skip connections from corresponding encoder layers, refines spatial accuracy, making U-Net especially effective in medical imaging and other precise segmentation applications. Both models are popular for transfer learning, adapting well to tasks with limited labeled data. The structures for each model is given in Figure \ref{fig:vgg_model} and \ref{fig:unet_model}.\\

\subsection{Transfer Learning}
\noindent Due to the high computational cost of training large models, transfer learning has become an essential part in deep learning. By initializing segmentation models with pre-trained weights that capture general features, models can quickly specialize in traffic-specific tasks like detecting lanes and signals. In road traffic segmentation, deep architectures like VGG16 and U-Net are fine-tuned by adjusting the final layers to classify each pixel in traffic images accurately.

\subsection{Loss Function and Optimizer}
\noindent For this task, Binary Cross-Entropy with Logits Loss (BCEWithLogitsLoss) is used since only two classes (road and background) are identified, coupled with the Adam optimizer due to its adaptive learning rate and efficiency in handling sparse gradients.

\subsection{Evaluation Metrics}
\subsubsection{Pixel Accuracy}
Pixel accuracy measures the percentage of correctly classified pixels across the entire image. While simple to compute, pixel accuracy may not be ideal for unbalanced datasets, as it tends to overrepresent dominant classes (e.g., background pixels in road images).
\subsubsection{Intersection over Union}
Intersection over Union (IoU), also known as the Jaccard Index, is calculated as the ratio of the intersection and union of predicted and actual segmentation masks for each class. IoU is a robust metric for segmentation as it evaluates spatial overlap, making it more sensitive to accurate boundary predictions. The mean IoU (mIoU) score is often used as an aggregate measure of performance across multiple classes.
\subsubsection{Precision and Recall}
From the confusion matrix, two key metrics—precision and recall—can be derived. These metrics are crucial for assessing the model’s performance, particularly when some classes (e.g., pedestrians or traffic signs) might be rarer or more critical than others. Precision measures how many of the pixels that the model predicted as a specific class are actually in that class. Recall, or sensitivity, measures how many of the actual pixels in a specific class are correctly identified by the model.
\subsubsection{F1 Score}
The F1 score combines precision and recall to evaluate class-specific performance, particularly useful for unbalanced datasets. 

\section{Methodology}
\noindent Before training, the models need to be modified sufficiently in order to perform the task at hand. The VGG-16 Model is modified by removing the fully connected layers at the end and replacing them with convolutional layers for upsampling, giving the model the ability to classify each pixel for image segmentation while allowing for transfer learning of earlier layers. The convolutional layers added are as follows:\\

\begin{minipage}{0.45\textwidth}
    \begin{itemize}[left=0pt]
        \item \textbf{ConvTranspose2d}\((\text{(512, 256)})\)
        \item \textbf{ReLU} \((\text{inplace=True})\)
        \item \textbf{ConvTranspose2d}\((\text{(256, 128)})\)
        \item \textbf{ReLU} \((\text{inplace=True})\)
        \item \textbf{ConvTranspose2d}\((\text{(128, 64)})\)
        \item \textbf{ReLU} \((\text{inplace=True})\)
        \item \textbf{Conv2d} \((\text{(64, 1)})\)
        \item \textbf{Upsample} \((\text{(512, 512)})\)
    \end{itemize}
\end{minipage}
\\\\
The U-Net model is modified by changing the output segmentation map to match the output resolution needed. No pre-trained weights are used for this model. \\\\
After modification, both models will be trained with separate datasets, selected at random:

\begin{itemize}
    \item \href{https://www.cvlibs.net/datasets/kitti/eval_road.php}{KITTI Road Segmentation: 289 Images}
    \item \href{https://github.com/commaai/comma10k}{Comma10k: 5,000 Images}
\end{itemize}

\begin{figure}[ht]
  \centering
  \fbox{\includegraphics[width=0.4\textwidth]{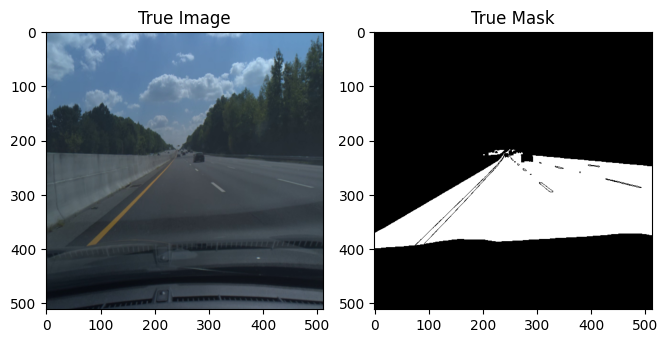}}
  \caption{An example of image segmentation from the Comma10k dataset. The black areas represent values of 0 while the white areas represent 1.}
  \label{fig:comma_seg_example}
\end{figure}

\noindent To prepare data for model training, we first split the data into a train/val/test split of (70\%/15\%/15\%). Subsequently, DataLoader and Dataset classes are created to transform the data into a compatible format. For our binary segmentation task, the segmentation data needs to be a tensor where each pixel is either 1 (for road) or 0 (for non-road). The datasets are initially in image format, where each class is represented by a specific color. We filtered out the color representing the road class, setting those pixels to 1 and all others to 0. This effectively reshapes the tensors from (3*W*H) to (1*W*H). An example of this can be seen in Figure \ref{fig:comma_seg_example}\\

\begin{figure}[ht]
  \centering
  \fbox{\includegraphics[width=0.4\textwidth]{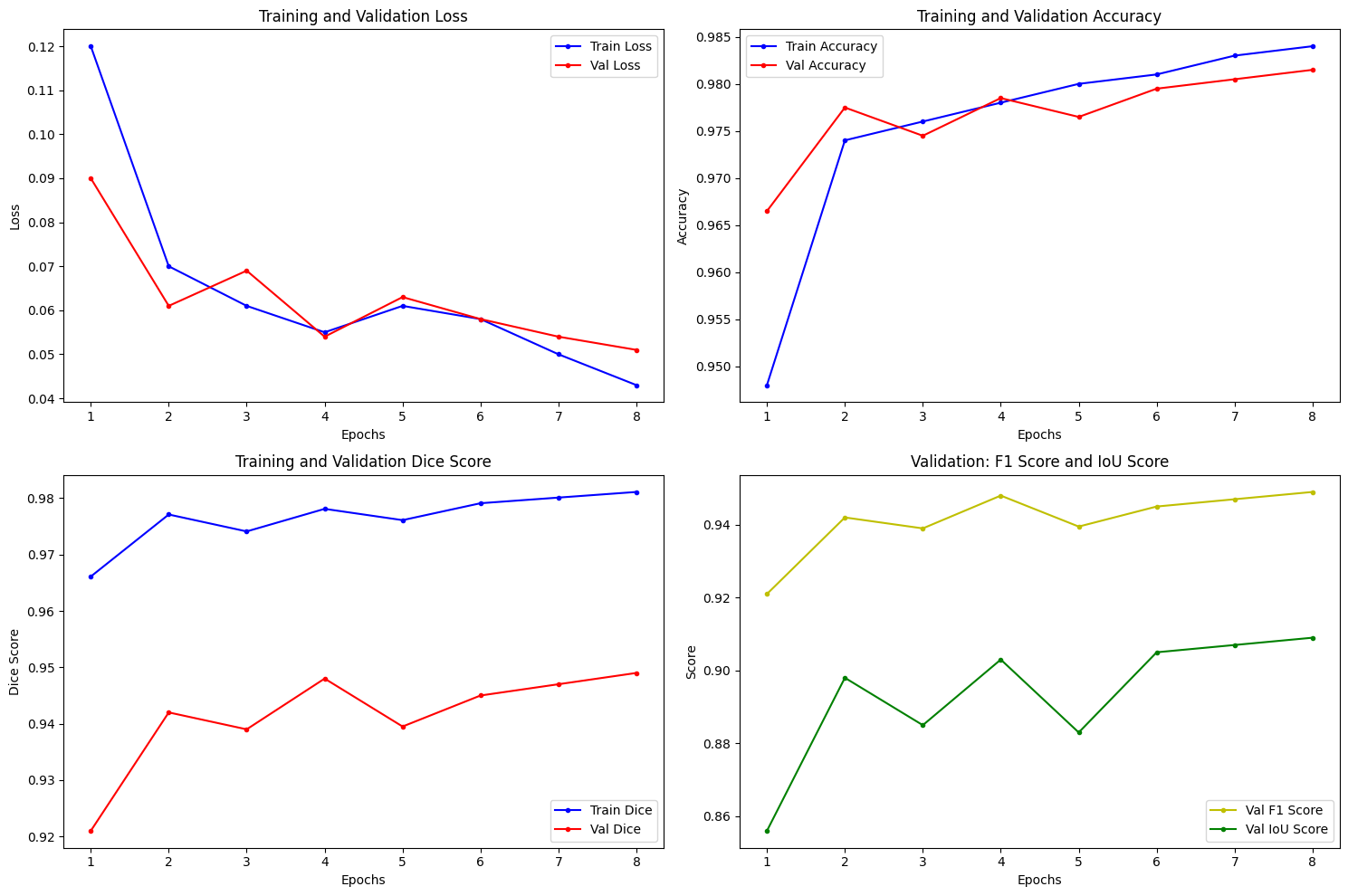}}
  \caption{Training and evaluation data after training the modified VGG-16 model on the Comma10k dataset for 8 epochs.}
  \label{fig:comma_training}
\end{figure}

\begin{figure}[ht]
  \centering
  \fbox{\includegraphics[width=0.4\textwidth]{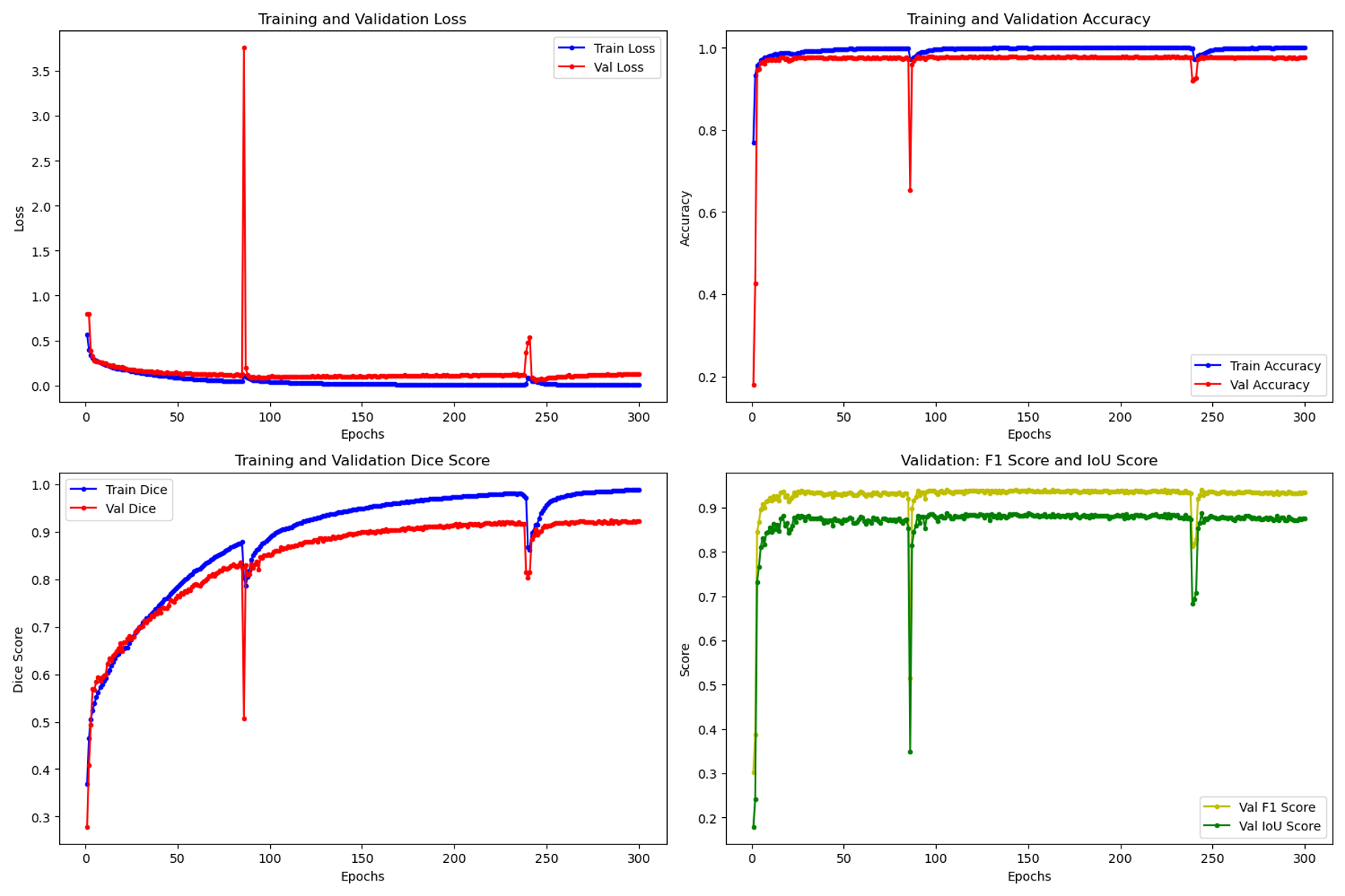}}
  \caption{Training and evaluation data after training the modified U-Net model on the KITTI dataset for 300 epochs.}
  \label{fig:kitti_training}
\end{figure}

\noindent Once the DataLoaders are set up, model training can begin. To ensure the models learn the data well, we trained to reach over 97\% pixel accuracy. The modified U-Net model achieved this accuracy in 300 epochs on the KITTI Road dataset, while the modified VGG-16 model reached it in just 8 epochs. Both sets of training graphs can be seen in Figures \ref{fig:comma_training} and \ref{fig:kitti_training}.

\section{Cross-Validation and Discussion}

\noindent Once the models are trained, cross-validation is performed by evaluating a model with the other dataset that it has not been trained on: 

\begin{itemize}
    \item Modified VGG-16 on KITTI 
    \item Modified U-Net on Comma10k
\end{itemize}

\begin{figure}[ht]
  \centering
  \fbox{\includegraphics[width=0.45\textwidth]{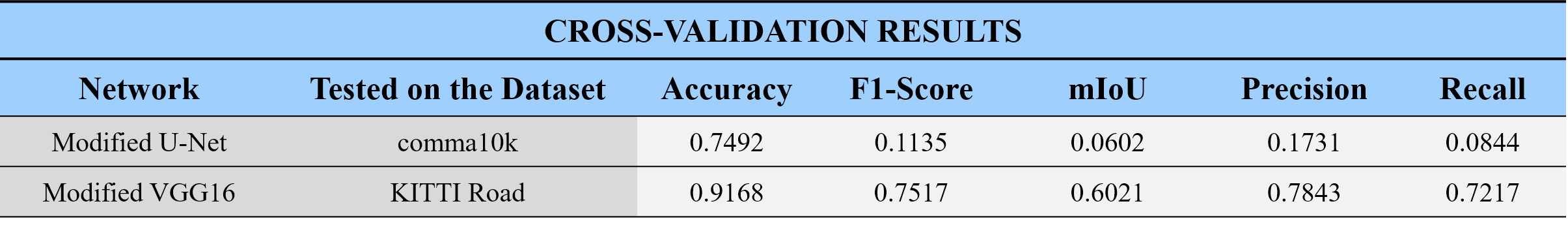}}
  \caption{Results after the cross-validation of models.}
  \label{fig:cross_validation_results}
\end{figure}

\noindent In this step, the training loop is completely removed from the code and only the validation loop is left. Validation is then performed for the entire dataset for one epoch, then the results are tabulated in Figure \ref{fig:cross_validation_results}.\\

\noindent The modified U-Net model underperformed significantly compared to the modified VGG-16 model, despite being trained with significantly more epochs. Additionally, U-Net scored lower across all other metrics. This suggests that the VGG-16 model effectively transferred knowledge gained from the Comma10k dataset to other datasets, like KITTI, whereas the modified U-Net model struggled to generalize its learning from the KITTI dataset. To investigate this, a deep dive is needed into dataset characteristics and model features.
\subsection{Dataset Size}
\noindent One of the most significant and obvious discrepancies between training both models would be the size of the dataset, the Comma10k dataset being 17 times bigger. When training a model on a small dataset, it often learns fine details and noise, rather than the general patterns, leading to overfitting. The best way to tackle this would be to use a combination of different datasets, where its combined diversity is high enough to improve generalization.
\subsection{U-Net Trained From Scratch}
The modified U-Net model was also not pre-trained, while the modified VGG-16 model already was. This means that the modified U-Net model was trained completely with the KITTI dataset, worsening its disability to generalize what it has learned.
\subsection{Dataset Characteristics}
\noindent Upon inspecting the datasets, it appears that the KITTI dataset mostly contains images of a suburban environment, the city of Karlsruhe, Germany, with narrow roads and many trees. The Comma10k dataset is taken from the California Highway, with wider roads and rarely any trees seen. The dataset at first seems worlds apart, but there are several reasons why the KITTI dataset performed worse:\\
\subsubsection{Tree Shadows}
\begin{figure}[ht]
  \centering
  \fbox{\includegraphics[width=0.45\textwidth]{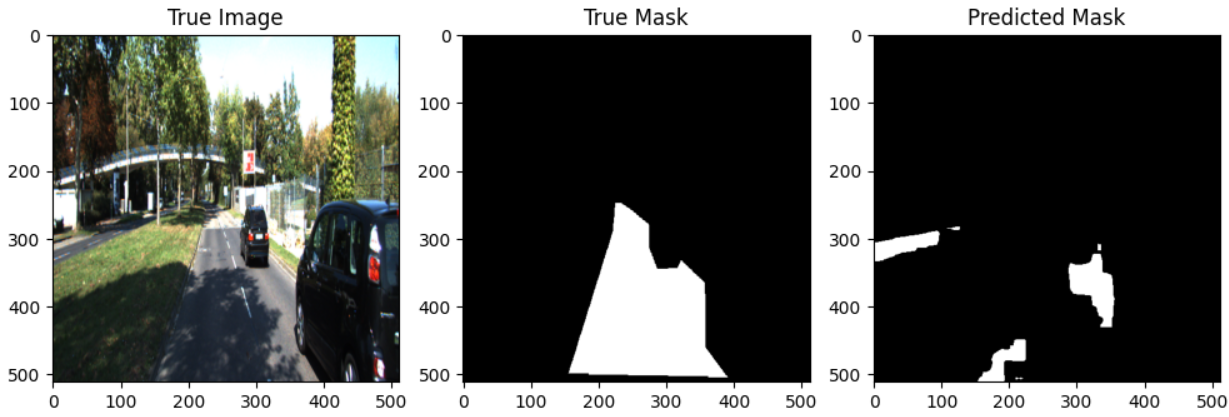}}
  \caption{The dark shadow of the tree prevents effective classification, the model being only able to classify the unshaded part of the road.}
  \label{fig:shadow}
\end{figure}

Due to the presence of many trees, the shadows are apparent enough and large enough to cast very dark shadows on the road, effectively obstructing the road almost completely from the camera. As we can see in Figure \ref{fig:shadow}.\\

\subsubsection{Pavements}
\begin{figure}[ht]
  \centering
  \fbox{\includegraphics[width=0.45\textwidth]{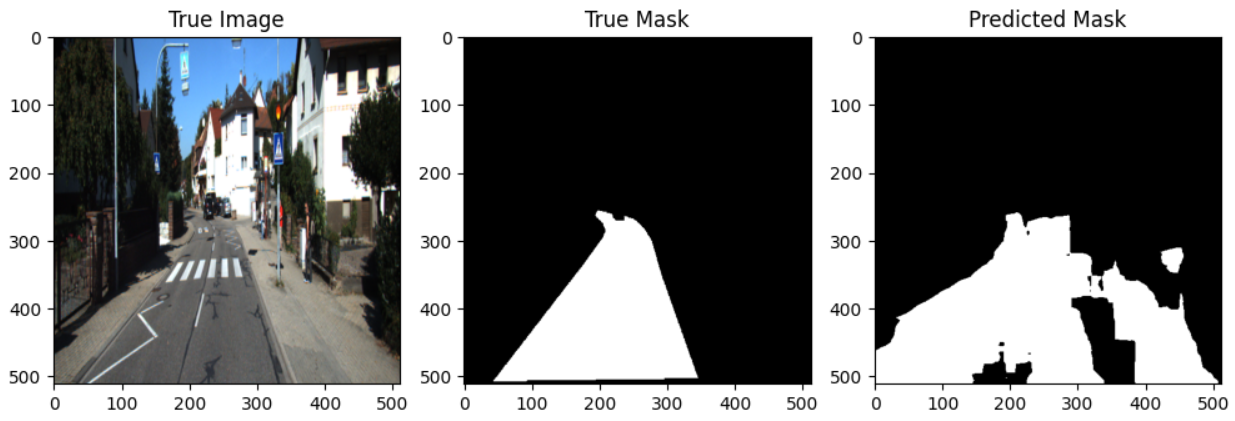}}
  \caption{The model misclassifies parts of the pavement as the road, due to their similarity.}
  \label{fig:pavement}
\end{figure}

The sidewalks have a similar color to the roads. This leads to constant misclassification of sidewalks as roads, as shown in Figure \ref{fig:pavement}\\

\subsection{Annotation Artifacts}
\begin{figure}[ht]
  \centering
  \fbox{\includegraphics[width=0.45\textwidth]{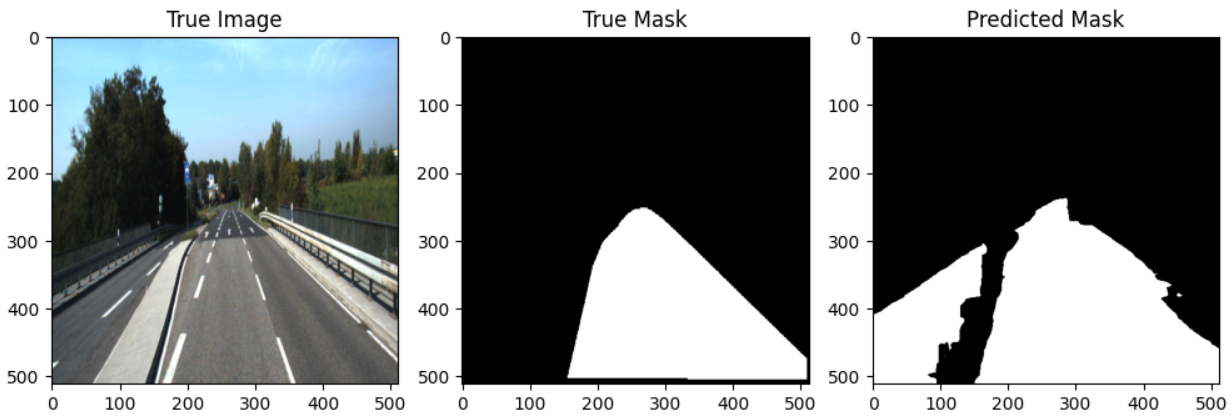}}
  \caption{The road on the left is unlabelled, causing the model to misclassify the left road as a road, even though it is correctly classified.}
  \label{fig:unlabelled_road}
\end{figure}

In the KITTI dataset, roads that are going in the opposite direction of the current vehicle are not annotated at all, whereas in the Comma10k dataset, roads facing the opposite way are usually not visible at all, since the roads are so wide. This means that the model trained with the Comma10k dataset is not trained on road directions and classifies every road that it sees. This discrepancy is shown in Figure \ref{fig:unlabelled_road}.

\subsubsection{Data Augmentation}

Lane markings in the Comma10k dataset were misannotated, creating boundary artifacts classified as non-road class. To fix this, we can apply augmentation techniques like morphological dilation to the lane marking mask, allowing it to merge with the road class and eliminate edge issues, ensuring accurate segmentation.

\begin{figure}[ht]
  \centering
  \fbox{\includegraphics[width=0.45\textwidth]{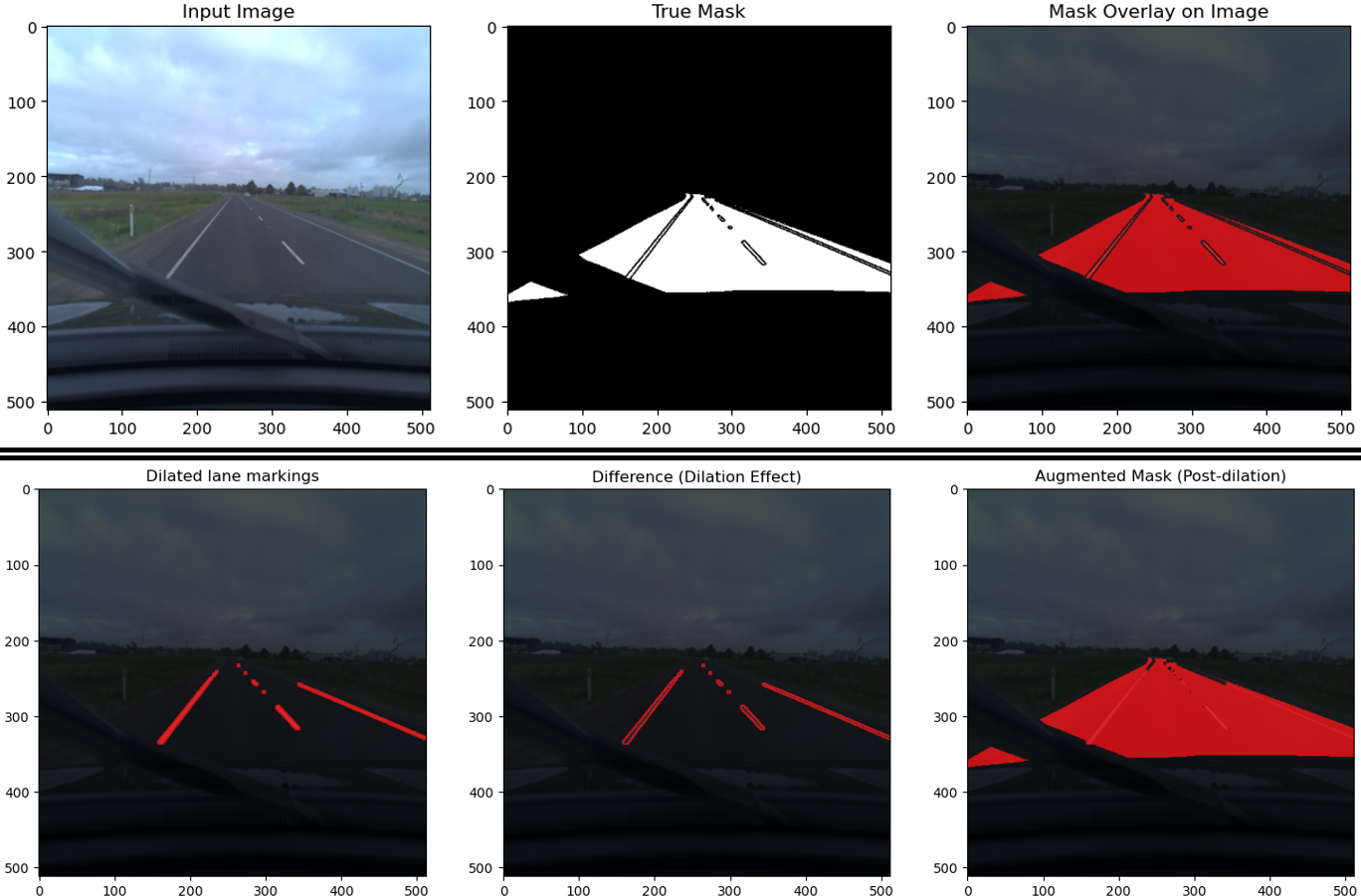}}
  \caption{Data augmentation techniques to remove misannotations, such as patches of roads annotated as non-road.}
  \label{fig:augmented_road}
\end{figure}
 
\noindent The aforementioned discrepancies in the dataset characteristics give rise to misclassification problems, which can be solved by choosing a different dataset.

\section{Conclusion}
\noindent This project demonstrates that the choice of dataset and model characteristics greatly affect the ability of a model to classify unseen data, as the datasets used for testing may have discrepancies that other datasets may not have, such as the presence of very dark tree shadows in the case of the KITTI dataset. Data augmentation techniques and transfer learning can be implemented to overcome the effects of said discrepancies. All in all, one should always explore and utilize different datasets to seek what is best for the project at hand, and seek datasets with a wide variation of data to reduce overfitting and improve generalization performance.

\end{document}